# In-depth analysis of recall initiators of medical devices with a Machine Learning-Natural language Processing tool

Yang Hu[1][a], Pezhman Ghadimi[1][b]

[1]*Laboratory for Advanced Manufacturing Simulation and Robotics, School of Mechanical & Materials Engineering*
*University College Dublin, Belfield, Dublin, Ireland*
*Yang.hu@ucdconnect.ie, pezhman.ghadimi@ucd.ie*

Keywords: Recall Initiator Analysis, Medical Device, Artificial Intelligence, Supply Chain Risk Management, Reverse Logistics

Abstract: Persistent quality problems with medical devices and the associated recall present potential health risks to patients and users, bringing extra costs to manufacturers and disturbances to the entire supply chain (SC). Recall initiator identification and assessment are the preliminary steps to prevent medical device recall. Conventional analysis tools are inappropriate for processing massive and multi-formatted data comprehensively and completely to meet the higher expectations of delicacy management with the increasing overall data volume and textual data format. To address these problems, this study presents a bigdata-analytics-based Machine learning (ML) – Natural language Processing (NLP) tool to identify, assess and analyse the medical device recall initiators based on the public medical device recall database from 2018 to 2024. Results suggest that the Density-Based Spatial Clustering of Applications with Noise (DBSCAN) clustering algorithm can present each single recall initiator in a specific manner, therefore helping practitioners to identify the recall reasons comprehensively and completely. This is then followed by text similarity-based textual classification to assist practitioners in controlling the group size of recall initiators and provide managerial insights from the operational to the tactical and strategic levels. More proactive practices and control solutions for medical device recalls are expected in the future.

## 1 INTRODUCTION

Medical devices play an increasingly significant role in healthcare delivery (Thirumalai & Sinha, 2011) which is especially witnessed after the global pandemic. The medical device industry has grown remarkably in revenues and technological sophistication (Sarkissian, 2018). However, several hundred medical device recalls occur each year (Gagliardi et al., 2017). In 2022, the U.S. Food and Drug Administration (FDA) reported 70 highest-risk recalls, compared to an average of 47 over the previous five years (Taylor, 2023). Serious medical device adverse events have overtaken industry growth by 8% and recalls have increased on par with the growth rate (Sarkissian, 2018). Persistent quality problems with medical devices and the associated recalls will not only present potential health risks to patients and personnel users of these devices (Mukherjee & Sinha, 2018; Thirumalai & Sinha, 2011) but also result in high extra costs to the manufacturer, its supply chain members (Ahsan & Gunawan, 2014; Morgenthaler et al., 2022) and the healthcare system (Ghobadi et al., 2019). Besides the huge litigation fees incurred, the recall event can result in estimated losses of billions of dollars in lost sales (Marucheck et al., 2011), as current and potential clients will turn to other competitors because of lost reputation (Blom & Niemann, 2022). Although medical devices have become an indispensable component (often lifesaving) in health care delivery, sometimes they also become sources of significant risk to the consumers of medical devices (Thirumalai & Sinha, 2011).

Recalls are reverse logistics where recalled products, information, and cash flow are in the opposite direction of the normal supply chain. The process of product recall is cumbersome and members of the entire supply chain are directly or indirectly affected by recalls (Ahsan & Gunawan, 2014).

---

[a]  https://orcid.org/0000-0001-5146-3572
[b]  https://orcid.org/0000-0003-0153-9035

A medical device recall is a voluntary action by manufacturers to remove or correct devices that violate the National Food and Drug Administration (e.g. FDA in the U.S, EMA in the EU or Health Canada in the CA) regulations. These corrections are often related to manufacturing defects, functional defects (Thirumalai & Sinha, 2011), software failure (Bliznakov, Mitalas, & Pallikarakis, 2007; Fu et al., 2017; Wallace & Kuhn, 2001), device design, process control, deceptive presentations and labelling (Sarkissian, 2018).

The FDA classifies medical devices according to the risk degree of health threat to users, from low (Class I) to high (Class III). Class I devices are deemed to pose the least amount of risk to patients since their designs are straightforward, they are easy to produce, and they do not pose any danger to patients, while Class III devices are of substantial importance in human health (Sarkissian, 2018) require a premarket approval application (PMA) (FDA, 2024a). The recall classification is the reverse of the medical device classification logic whereby class III recalls only reflect regulatory volitions with minimal or no health risks, Class I recalls denote situations in which exposure to a product will cause serious adverse health consequences or death (Sarkissian, 2018). For a Class I recall, the company will notify the customers and issue a press release to notify the public (Villarraga, Guerin, & Lam, 2007).

Medical device recalls are not uncommon, and the safety of medical devices may pose public health risks (Gagliardi et al., 2017). Recall initiator identification, assessment, and analysis are used to overcome and prepare for a possible medical product recall (Ahsan & Gunawan, 2014). Recall risk factor identification and assessment are the preliminary steps to prevent medical device recall. Controlling the recall operations of medical devices and figuring out high-risk products to strengthen anticipatory risk control action can improve the quality of the utilisation for users and reduce recall operations.

Realizing the importance of medical device recalls (Ahsan & Gunawan, 2014; Gagliardi et al., 2017; Villarraga et al., 2007), researchers analysed medical recall initiators with historical recall data from different time periods to provide insights into recall trends (Ahsan & Gunawan, 2014; Gagliardi et al., 2017; Sarkissian, 2018) with commonly used data analysis tools. However, with the increasing data volume and widely used textual data, conventional analysis tools are not appropriate for processing massive and muti-formatted data comprehensively and completely to meet the higher expectations of dealing efficiency and delicacy management.

To address the shortcomings in dealing efficiency and data process versatility of conventional tools in the practical context of big data volume and muti data format, this study presents a Machine Learning–Natural Language Processing work tool based on big data analytics that remained unexplored by previous studies to identify and analyse the medical device recall initiators in a comprehensive and complete manner and to present up-to-date information concerning medical device recalls according to the publicly available FDA medical device recall database from 2018 to 2024.

This information contributes to the literature on the risk identification and assessment of medical device supply chain. It is also relevant to policymakers, health system leaders, clinicians, and regulators to understand the possible public health risks posed by medical devices and to explore whether current approaches to post-market surveillance of medical devices are appropriate.

This research also contributes a new analytical work tool for the supply chain risk analysis research community to approach the goal of efficiency, reliability, and thoroughness. From the digital technology implementation perspective, this research manage to expand the application scenarios of AI to the reverse side of medical device supply chain that is neglected by previous studies (Hu & Ghadimi, 2023).

The paper is organized as follows: in Section 2, a literature review of the medical device recall analysis and previous relevant research are discussed. Section 3 describes the data collection and research methodology process. This is followed by results and discussion in Section 4. Finally, the research summary and future research directions are highlighted in the conclusion in Section 5.

## 2 LITRETURE REVIEW

Table 1. illustrates previous research on medical device recall analysis that was categorized by recall product category, recall initiator category, data analysis tool, analysis period and data source.

Table 1: Literature Categorization.

| R | Device | Recall Initiator | Analysis Tool | Analysis Period | Dataset |
|---|---|---|---|---|---|
| (Wallace & Kuhn, 2001) | Anesthesiology, cardiology, diagnostics, radiology, general hospital use, and surgery categories | Software failure | Conventional | 1983-1997 | FDA |
| (Bliznakov et al., 2007) | Device class I-III | Software failure | Conventional | 1999-2005 | FDA |
| (Villarraga et al., 2007) | Device class I-III | Recall class I | Conventional | 2004-2006 | FDA |
| (Yi, Shenglin, Qiang, & Hanxi, 2013) | Device class III | Recall class I-III | Conventional | 2005-2006 | FDA, US. |
| (Somberg, McEwen, & Molnar, 2014) | Cardiovascular and Noncardiovascular ccategpry | Recall class I-III | Conventional | 2005-2012 | FDA |
| (Connor et al., 2017) | Radiation Oncology Category | Recall class I-III | Conventional | 2002-2015 | FDA |
| (Gagliardi et al., 2017) | Device class I-IV | Recall Class I-III | Conventional | 2005-2015 | Health Canada, |
| (Sarkissian, 2018) | Device class III | Recall Class I | Conventional | 2014-2018 | FDA. |
| (Vajapey & Li, 2020) | Orthopaedics category | Recall class I-II | Conventional (Excel) | 2015-2019 | FDA |
| Present study | Device class I-III | Recall class I-III | Big Data and AI | 2018-2024 | FDA |

The previous analysis investigated partly either recall device category or recall initiators. Gagliardi et al. (2017) performed a comprehensive analysis that went through all types of devices and recall initiators in the Canadian region wide. However, the comprehensive analysis based on other geographical spaces is limited. The previous analyses do not reveal the root causes of medical device recalls, which is critical to help manufacturers understand the failures and prevent recalls in the future (Fu et al., 2017). Moreover, the analytic tools leveraged by previous studies are conventional tools such as Excel (Vajapey & Li, 2020). However, with the increasing data volume, and widely used textual data, conventional analysis tools are not appropriate for processing massive, muti-formatted data and meeting the higher expectations of dealing with efficiency (Sagiroglu & Sinanc, 2013) and delicacy management. Lastly, the information leveraged by previous research is out of date, the medical device recall analysis based on nearly 5 years has not been presented.

To address these gaps, this research proposed a Machine Learning – Natural Language work tool based on big data analytics that remained unexplored by previous studies to identify and analyse the medical device recall initiators, presenting up-to-date information concerning medical device recalls according to the public medical device recall database from 2018 to 2024. This research explores a new attempt at medical device recall initiator analysis from a methodology perspective. This research attempts to overcome the shortcomings in dealing efficiency and data process versatility of conventional tools in the practical context of big data volume and muti data format with AI tool.

## 3 DATA COLLECTION AND RESEARCH METHODOLOGY

### 3.1 Data Collection

Data is scraped from the FDA open database of medical device recall using API calls (FDA, 2024c). The FDA database only allowed 1000 records to be retrieved at once, the maximum loop tested successfully by the current computer device is 7. Therefore, 7000 rows of data records were included in the final version dataset used for data analysis of this research, dated from January 1, 2018, to April 15, 2024. This research organized an information profile for analysis that includes contents: 1) product code, 2) recall posted date, 3) recalling firm, 4) root cause description, 5) product quantity, 6) device name and 7) device class by using two API addresses.

Columns 1-5 were extracted from URL= https://api.fda.gov/device/recall.json while URL= https://api.fda.gov/device/classification.json is the source of columns 6-7. These two separate datasets can be merged and tailored to the one that meet the requirements for this research as they share the same 'product_code' column. The merged dataset for recall initiators analysis in this research is briefly illustrated in Fig 1.

Figure 1: Scrapped data by API application.

In Fig 1, it can be found that the root cause description contains human-written, unstructured (Fu et al., 2017) short text determining the general type of recall cause by FDA (FDA, 2024b) such as 'Process design, 'Nonconforming Material/Component', 'Under Investigation by firm' 'Device design', 'Employee error', 'Process control' and 'Other'. This research considers the contents in the 'root_cause_description' column as the recall initiators. With the unstructured short text data, some classic analysis methods are not applicable. Machine learning is a capable tool for dealing with massive data with both numerical and short textual format (Sun, 2019).

Before implementing the machine learning approach, the merged dataset was cleaned by removing all special characters, null values, duplicated data, outliers, and any other content that does not add value to the analytics results. All the analysis work was performed on the Google Colab platform using Python 3.10.

## 3.2 Research Method

Machine Learning (ML) is defined as "the field of study interested in the development of computer algorithms to transform data into intelligent actions" (Sheridan et al., 2020). ML has been used in medical device-related research to identify and discover trends and patterns (Xu & Chan, 2019) to improve process (Kovačević, Gurbeta Pokvić, Spahić, & Badnjević, 2020; Raschka & Mirjalili, 2019; Xu & Chan, 2019). Machine learning algorithms can be categorized into three groups: supervised learning, unsupervised learning, and reinforcement learning (Raschka, 2015). The unsupervised learning techniques such as clustering can be used to discover hidden structures or patterns, grouping the separated data based on their similarities for the unstructured input short text data in the 'root cause description' column of this research. Clustering is the appropriate machine learning technique for this research as it provides a means for identifying trends and patterns that may not be obvious.

The Density-Based Spatial Clustering of Applications with Noise (DBSCAN) algorithm was leveraged to identify the recall initiators in this research. In face of 7000 or more records of recall initiators in this research, point out the cluster numbers ahead is difficult, the DBSCAN does not require users to select the number of clusters to use arbitrarily, compared to other commonly used classical clustering methods, such as Spectral Clustering and K-means clustering (Murugesan, Cho, & Tortora, 2021). Moreover, the DBSCAN can be used with both numerical data and short text data that used for recording the recall initiators and is suitable for identifying outliers (Sheridan et al., 2020). It requires the choice of two user-defined parameters, the neighbourhood distance epsilon (ε) and the minimum number of points (minPts) (Çelik, Dadaşer-Çelik, & Dokuz, 2011) or so-called MinSamples (Murugesan et al., 2021). The number of clusters is generated as a product of the analysis, and instances in low-density regions are tagged as outliers rather than assigned to a cluster. A cluster forms when there is at least a minPts within a user-specified threshold ε of a given point.

The small ε values often result in large numbers but small sizes of clusters (Sheridan et al., 2020) as the overlap probability of different clusters becomes small since the radius of each cluster is quite small.

The MinSamples represents the minimum number of points required to form a cluster, Smaller MinSamples generally results in large number clusters but small size of each cluster, as fewer points are required to form a dense region.

The small ε and small MinSamples can be helpful in this research to detect the recall initiators completely with the least duplicate or overlap.

This research setting ε=0.1, minPts = 4 based on the optimal results from (Murugesan et al., 2021). To avoid the possible bias, this research also tested the scenarios that MinPts= [1,2,3]. The cluster results remain the same with minPts =4.

## 4 RESUTLS AND DISCUSSION

Table 2 illustrates the results of recall initiator identification by the DBSCAN algorithm.

Table 2: Recall initiators identified after the DBSCAN clustering.

| Cluster | Recall initiator | Case Number |
|---|---|---|
| 1 | Other | 197 |
| 2 | No Marketing Application | 45 |
| 3 | Under Investigation by firm | 1699 |
| 4 | Software design | 270 |
| 5 | Radiation Control for Health and Safety Act | 43 |
| 6 | Material/Component Contamination | 42 |
| 7 | Device Design | 1046 |
| 8 | Employee error | 94 |
| 9 | Process control | 1030 |
| 10 | Process change control | 125 |
| 11 | Error in labelling | 98 |
| 12 | Software Manufacturing/Software Deployment | 13 |
| 13 | Component design/selection | 131 |
| 14 | Software Design Change | 45 |
| 15 | Labelling Change Control | 81 |
| 16 | Labelling design | 108 |
| 17 | Process design | 135 |
| 18 | Incorrect or no expiration date | 23 |
| 19 | Software change control | 16 |
| 20 | Mixed-up of materials/components | 29 |
| 21 | Component change control | 116 |
| 22 | Unknown/Undetermined by firm | 165 |
| 23 | Nonconforming Material/Component | 643 |
| 24 | Packaging | 49 |
| 25 | Labelling mix-ups | 34 |
| 26 | Packaging process control | 135 |
| 27 | Vendor change control | 99 |
| 28 | Storage | 134 |
| 29 | Equipment maintenance | 72 |
| 30 | Pending | 51 |
| 31 | Software design (manufacturing process) | 13 |
| 32 | Use error | 33 |
| 33 | Packaging change control | 49 |
| 34 | Package design/selection | 18 |
| 35 | Labelling False and Misleading | 14 |
| 36 | Environmental control | 96 |

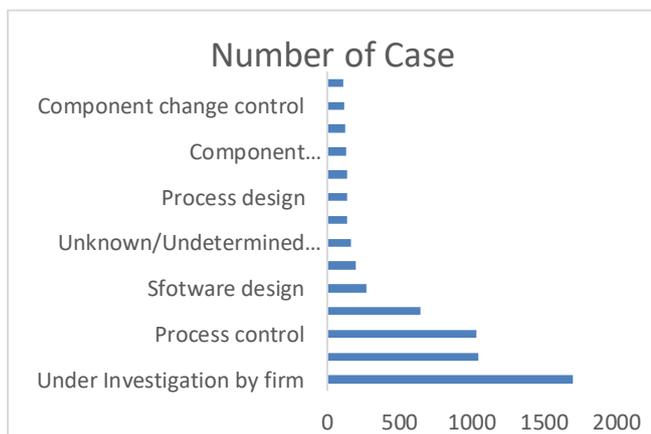

Figure 2: Rank of recall initiators after implementing the DBSCAN algorithm.

36 recall initiators are listed in Table 2 after the DBSCAN clustering. The most important reason for the medical device recall of the 7000 records is 'Under investigation by firm', accounting for 24.3 % of all the recall cases. This was followed by the 'Device design' reason that resulted in 1046 cases of medical device recall. 'Process control' is also an important recall initiator and ranked third place among all recall initiators. The DBSCAN clustering presents each single recall initiator in a specific manner and can help practitioners identify the recall reasons comprehensively and completely.

And if we check Table 2 closely, it can be noticed that some of these listed recall initiators can be aggregated and displayed under the same label (Sarkissian, 2018).

For instance, cluster 4 'Software design', cluster 12 'Software Manufacturing/Software Deployment', cluster 14 'Software Design Change', cluster 19 'Software change control' and cluster 31 'Software design (manufacturing process)' can be aggregated in a same 'Software' label (Connor et al., 2017). Similar situations such as cluster 9,10,17 can be aggregated in a 'Process' label, cluster 15,16,25, and 35 can be aggregated in a 'Labelling' label (Sarkissian, 2018).

The entire list of 36 exact recall initiators retrieved by the DBSCAN algorithm is useful for practitioners at the operational level, while the aggregated label can release the burden of investigating every specific detail for the practitioners at the tactical and strategical levels. In this case, to make the results of the recall initiator identification more widely acceptable, this research presents a textual classification step to aggregate the label based on the text similarity after the DBSCAN clustering. This is a new attempt to aggregate recall reasons using Natural Language Processing (NLP) techniques, compared to previous studies (Sarkissian, 2018) that rely on manual observation and experience (Connor et al., 2017). The computer-directed NLP techniques can help reduce manual work and efficiently perform in large datasets.

This research leveraged the text similarity measure for cluster aggregate with the first 10 letters of each recall reason phrase (Kenter & De Rijke, 2015). The results of combining groups by text similarity are presented in Table 3.

Table 3: Results of recall initiators clustering after textual classification.

| Cluster | Recall initiator | Case Number |
|---|---|---|
| 1 | ['Component change control', 'Component design/selection'] | 247 |
| 2 | ['Device Design'] | 1046 |
| 3 | ['Employee error'] | 94 |
| 4 | ['Environmental control'] | 96 |
| 5 | ['Equipment maintenance'] | 72 |

| 6 | ['Error in labelling'] | 98 |
|---|---|---|
| 7 | ['Incorrect or no expiration date'] | 23 |
| 8 | ['Labelling Change Control', 'Labelling design', 'Labelling False and Misleading', 'Labelling mix-ups'] | 237 |
| 9 | ['Material/Component Contamination'] | 42 |
| 10 | ['Mixed-up of materials/components'] | 29 |
| 11 | ['No Marketing Application'] | 45 |
| 12 | ['Nonconforming Material/Component'] | 643 |
| 13 | ['Other'] | 197 |
| 14 | ['Package design/selection', 'Process design'] | 153 |
| 15 | ['Packaging', 'Packaging change control', 'Packaging process control'] | 233 |
| 16 | ['Pending'] | 51 |
| 17 | ['Process change control', 'Process control'] | 1155 |
| 18 | ['Radiation Control for Health and Safety Act'] | 43 |
| 19 | ['Software design'] | 270 |
| 20 | ['Software change control', 'Software design (manufacturing process)', 'Software Design Change', 'Software Manufacturing/Software Deployment'] | 87 |
| 21 | ['Storage'] | 134 |
| 22 | ['Under Investigation by firm'] | 1699 |
| 23 | ['Unknown/Undetermined by firm'] | 165 |
| 24 | ['Use error'] | 33 |
| 25 | ['Vendor change control'] | 99 |

Results in Table 3 indicate that the number of clusters decreased to 25 from 36 after the labelling of the group aggregate. The textual classification step after the DBSCAN clustering can assist users in controlling the group size of recall initiators to gain insights beyond the operational level.

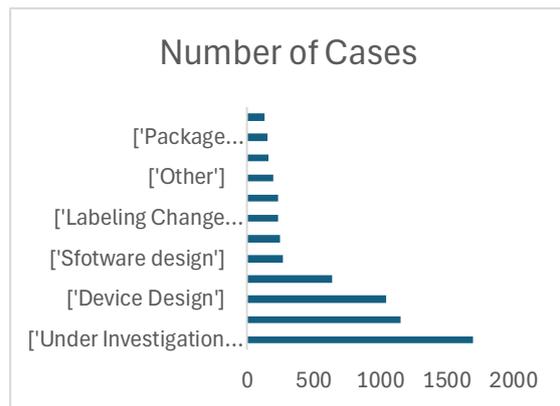

Figure 3: Rank of recall initiators after implementing the DBSCAN algorithm.

Table 4: Top 10 recall initiators comparison after Natural Language Process work.

| Top 10 Recall Reasons Before NLP | Top 10 Recall Reasons After NLP |
|---|---|
| Under investigation by firm | ['Under Investigation by firm'] |
| Device Design | ['Process change control', 'Process control'] |
| Process control | ['Device Design'] |
| Nonconforming Material/Component | ['Nonconforming Material/Component'] |
| Software design | ['Software design'] |
| Other | ['Component change control', 'Component design/selection'] |
| Unknown/Undetermined by firm | ['Labelling Change Control', 'Labelling design', 'Labelling False and Misleading', 'Labelling mix-ups'] |
| Packaging process control | ['Packaging', 'Packaging change control', 'Packaging process control'] |
| Process design | ['Other'] |
| Storage | ['Unknown/Undetermined by firm'] |

From the comparison of the results in Table 4, it can be found that the high-risk factors of medical device recall were identified differently after the textual classification. If the users determine the prior recall initiators based on the exact results from the DBSCAN, it might be misleading in some circumstances. For example, 'Device Design' is recognized as the second important recall initiator if we only apply DBSCAN. However, after the textual classification and label aggregate, it has been found that the 'Process control' problem should be the second priority and medical device manufacturers need to place enough resources to address it for fewer recall events with better performance. This ML-NLP work tool can not only capture specific details of each recall initiator, but also interpret the inner connection of each existing initiator that is recorded manually, and finally present a reasonable result that is closer to practical for users.

Besides the recall initiator identification and assessment supported by the ML-NLP work tool illustrated above, other results that may raise proper user interest can also be clearly stated with the assistance of big data analytics, such as the top recalled manufactures for the public attention in and top recalled products for manufactures to develop a safer production process to control recall risk of medical device.

The discussion above demonstrates that this big-data analytics-based ML-NLP tool can replicate the functions and results of conventional analysis tools as noted in previous studies. Additionally, Tables 2-4 and Figures 2-3 show that, beyond numerical analysis, this AI-supported tool can identify key features of medical device recalls from unstructured text, a task that typically required more manual effort with traditional tools. This ML-NLP tool captures features in a more automated, intelligent, and efficient manner, with fewer data format limitations compared to conventional analysis methods. The advantages in data processing versatility and efficiency become more pronounced as the data volume increases. The faster processing speed enables medical manufacturers to quickly identify and assess recall initiators, ultimately leading to faster implementation of risk prevention measures. This can help reduce public health risks and lower additional costs for the entire medical device system.

## 5 CONCLUSITONS

The big data analytics and artificial intelligence is neglected by previous medical device initiators analysis studies leads that the root causes of medical device recalls are not revealed comprehensively and completely. However, being able to identify the root causes of the failures in depth is critical to help manufactures understand the failures and prevent recalls in the future (Fu et al., 2017) .

This research contributed a ML-NLP work tool based on the bigdata analytics techniques to identify, assess and analyse the recall initiators of medical devices, with comprehensiveness and practical reasonable considerations, this research presented up-to-date information concerning medical device recalls.

Not limited to recall initiator analysis, the bigdata analytics-based ML-NLP work tool proposed by this research can be leveraged as a scalable solution for general scenarios in the domain of risk identification and risk assessment in both forward and reverse side of supply chain. This research contributes a new risk analysis work tool for the supply chain risk management community.

Further research can be (1) applying this ML-NLP risk analysis work tool to other industry domain and in the forward side of supply chain (2) implementing synonym analysis in this ML-NLP work tool. In natural language, the authors use various expressions to express the same opinions (Sun, 2019). For example, 'Under Investigation by firm' and 'Unknown/Underdetermined by firm' in Table 4 can be recognized in the same cluster by the meaning of natural language with synonym analysis. Moreover, the choice of two parameters of DBSCAN might be difficult, especially for the inexperienced user of this ML algorithms, therefore, a new method for automatic determining of DBSCAN parameters (Starczewski, Goetzen, & Er, 2020) can be proposed and embedded in this ML-NLP work tool to make this work tool more intelligent and user friendly would be an interesting attempt in the future.

Investigating the consequence of medical device recall, exploring how to use information about recalls (Gagliardi et al., 2017) and developing more effective preventive and control solutions (Thirumalai & Sinha, 2011) to reduce to medical device recall are also eligible research path in the future.

## REFERENCES

Ahsan, K., & Gunawan, I. (2014). Analysis of product recalls: Identification of recall initiators and causes of recall. Operations and Supply Chain Management: An International Journal, 7(3), 97-106.

Bliznakov, Z., Mitalas, G., & Pallikarakis, N. (2007). Analysis and classification of medical device recalls. Paper presented at the World Congress on Medical Physics and Biomedical Engineering 2006: August 27–September 1, 2006 COEX Seoul, Korea "Imaging the Future Medicine".

Blom, T., & Niemann, W. (2022). Managing reputational risk during supply chain disruption recovery: A triadic logistics outsourcing perspective.

Çelik, M., Dadaşer-Çelik, F., & Dokuz, A. Ş. (2011). Anomaly detection in temperature data using DBSCAN algorithm. Paper presented at the 2011 international symposium on innovations in intelligent systems and applications.

Connor, M. J., Tringale, K., Moiseenko, V., Marshall, D. C., Moore, K., Cervino, L., . . . Pawlicki, T. (2017). Medical device recalls in radiation oncology: analysis of US Food and Drug Administration data, 2002-2015. International Journal of Radiation Oncology* Biology* Physics, 98(2), 438-446.

FDA. (2024a). Classify Your Medical Device. Retrieved from https://www.fda.gov/medical-devices/overview-device-regulation/classify-your-medical-device

FDA. (2024b). Devicer Recall Reference. Retrieved from https://open.fda.gov/fields/devicerecall_reference.pdf

FDA. (2024c). Searchable Fields. Retrieved from https://open.fda.gov/apis/device/recall/searchable-fields/

Fu, Z., Guo, C., Zhang, Z., Ren, S., Jiang, Y., & Sha, L. (2017). Study of software-related causes in the FDA medical device recalls. Paper presented at the 2017 22nd International Conference on Engineering of Complex Computer Systems (ICECCS).

Gagliardi, A. R., Takata, J., Ducey, A., Lehoux, P., Ross, S., Trbovich, P. L., . . . Urbach, D. R. (2017). Medical device recalls in Canada from 2005 to 2015. International journal of technology assessment in health care, 33(6), 708-714.

Ghobadi, C. W., Janetos, T. M., Tsai, S., Beaumont, J. L., Welty, L., Walter, J. R., & Xu, S. (2019). Approval-adjusted recall rates of high-risk medical devices from 2002-2016 across Food and Drug Administration device categories. Issues L. & Med., 34, 77.

Hu, Y., & Ghadimi, P. (2023). A Review of Artificial Intelligence Application on Enhancing Resilience of Closed-loop Supply Chain. Paper presented at the 2023 IEEE International Conference on Engineering, Technology and Innovation (ICE/ITMC).

Kenter, T., & De Rijke, M. (2015). Short text similarity with word embeddings. Paper presented at the Proceedings of the 24th ACM international on conference on information and knowledge management.

Kovačević, Ž., Gurbeta Pokvić, L., Spahić, L., & Badnjević, A. (2020). Prediction of medical device performance using machine learning techniques: infant incubator case study. Health and Technology, 10(1), 151-155.

Morgenthaler, T. I., Linginfelter, E. A., Gay, P. C., Anderson, S. E., Herold, D., Brown, V., & Nienow, J. M. (2022). Rapid response to medical device recalls: an organized patient-centered team effort. Journal of Clinical Sleep Medicine, 18(2), 663-667.

Mukherjee, U. K., & Sinha, K. K. (2018). Product recall decisions in medical device supply chains: a big data analytic approach to evaluating judgment bias. Production and Operations Management, 27(10), 1816-1833.

Murugesan, N., Cho, I., & Tortora, C. (2021). Benchmarking in cluster analysis: a study on spectral clustering, DBSCAN, and K-Means. Paper presented at the Data Analysis and Rationality in a Complex World 16.

Raschka, S. (2015). Python machine learning: Packt publishing ltd.

Raschka, S., & Mirjalili, V. (2019). Python machine learning: Machine learning and deep learning with Python, scikit-learn, and TensorFlow 2: Packt publishing ltd.

Sagiroglu, S., & Sinanc, D. (2013). Big data: A review. Paper presented at the 2013 international conference on collaboration technologies and systems (CTS).

Sarkissian, A. (2018). An exploratory analysis of US FDA Class I medical device recalls: 2014–2018. Journal of medical engineering & technology, 42(8), 595-603.

Sheridan, K., Puranik, T. G., Mangortey, E., Pinon-Fischer, O. J., Kirby, M., & Mavris, D. N. (2020). An application of dbscan clustering for flight anomaly detection during the approach phase. Paper presented at the AIAA Scitech 2020 Forum.

Somberg, J. C., McEwen, P., & Molnar, J. (2014). Assessment of cardiovascular and noncardiovascular medical device recalls. The American Journal of Cardiology, 113(11), 1899-1903.

Starczewski, A., Goetzen, P., & Er, M. J. (2020). A new method for automatic determining of the DBSCAN parameters. Journal of Artificial Intelligence and Soft Computing Research, 10(3), 209-221.

Sun, H. (2019). Sourcing Risk Detection and Prediction with Online Public Data: An Application of Machine Learning Techniques in Supply Chain Risk Management: North Carolina State University.

Taylor, N. P. (2023). FDA Class I recalls hit 15-year high in 2022. Retrieved from https://www.medtechdive.com/news/fda-class-i-recall-2022-ABT-BAX-GEHC-MDT-PHG/644072/

Thirumalai, S., & Sinha, K. K. (2011). Product recalls in the medical device industry: An empirical exploration of the sources and financial consequences. Management science, 57(2), 376-392.

Vajapey, S. P., & Li, M. (2020). Medical device recalls in orthopedics: recent trends and areas for improvement. The Journal of arthroplasty, 35(8), 2259-2266.

Villarraga, M. L., Guerin, H. L., & Lam, T. (2007). Medical Device Recalls from 2004-2006: A Focus on Class I Recalls. Food & Drug LJ, 62, 581.

Wallace, D. R., & Kuhn, D. R. (2001). Failure modes in medical device software: an analysis of 15 years of recall data. International Journal of Reliability, Quality and Safety Engineering, 8(04), 351-371.

Xu, S., & Chan, H. K. (2019). Forecasting medical device demand with online search queries: A big data and machine learning approach. Procedia Manufacturing, 39, 32-39.

Yi, Y., Shenglin, L., Qiang, Z., & Hanxi, W. (2013). Analysis of medical device recall reports in FDA database in 2005-2006. Paper presented at the World Congress on Medical Physics and Biomedical Engineering May 26-31, 2012, Beijing, China.